\documentclass[10pt,twocolumn,letterpaper]{article}

\usepackage{iccv}
\usepackage{times}
\usepackage{epsfig}
\usepackage{graphicx}
\usepackage{amsmath}
\usepackage{amssymb}

\usepackage{capt-of}
\usepackage{pifont}

\usepackage{bbm}

\usepackage{capt-of}
\usepackage{booktabs}
\usepackage{pgffor}
\usepackage{pdfpages}

\usepackage[accsupp]{axessibility}  

\usepackage[breaklinks=true,bookmarks=false]{hyperref}

\iccvfinalcopy

\def\iccvPaperID{****} 

\renewcommand{\vec}[1]{\boldsymbol{#1}}
\DeclareMathOperator{\loss}{\mathcal{L}}
\DeclareMathOperator{\E}{\mathbb{E}}
\DeclareMathOperator{\reals}{\mathbb{R}}

\usepackage[capitalize]{cleveref}
\crefname{section}{Sec.}{Secs.}
\Crefname{section}{Section}{Sections}
\Crefname{table}{Table}{Tables}
\crefname{table}{Tab.}{Tabs.}

\begin{document}

\title{Unaligned 2D to 3D Translation with Conditional\\ Vector-Quantized Code Diffusion using Transformers}
\author{
\begin{tabular}{@{}c@{}}
    Abril Corona-Figueroa$^1$ \qquad
    Sam Bond-Taylor$^1$ \qquad
    Neelanjan Bhowmik$^1$ \qquad \\
    Yona Falinie A. Gaus$^1$ \qquad 
    Toby P. Breckon$^{1,2}$ \qquad 
    Hubert P. H. Shum$^1$ \qquad 
    Chris G. Willcocks$^1$ \qquad
\end{tabular}
\\Department of \{$^1$Computer Science $\mid$ $^2$Engineering\}, Durham University,
Durham, UK
}
\maketitle

\begin{abstract}
Generating 3D images of complex objects conditionally from a few 2D views is a difficult synthesis problem, compounded by issues such as domain gap and geometric misalignment. For instance, a unified framework such as Generative Adversarial Networks cannot achieve this unless they explicitly define both a domain-invariant and geometric-invariant joint latent distribution, whereas Neural Radiance Fields are generally unable to handle both issues as they optimize at the pixel level. By contrast, we propose a simple and novel 2D to 3D synthesis approach based on conditional diffusion with vector-quantized codes. 
Operating in an information-rich code space enables high-resolution 3D synthesis via full-coverage attention across the views. Specifically, we generate the 3D codes (e.g. for CT images) conditional on previously generated 3D codes and the entire codebook of two 2D views (e.g. 2D X-rays). Qualitative and quantitative results demonstrate state-of-the-art performance over specialized methods across varied evaluation criteria, including fidelity metrics such as density, coverage, and distortion metrics for two complex volumetric imagery datasets from in real-world scenarios.
\end{abstract}

\section{Introduction} \label{s:intro}

\begin{figure}
\centering
\includegraphics[width=\columnwidth]{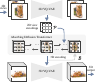}
\vspace{5pt}\captionof{figure}{The proposed approach independently learns a discrete information-rich codebook for 2D and 3D domains with two VQ-VAEs. New 3D samples are then synthesized conditional on the complete codebook of multiple 2D views with a transformer.}
\label{fig:task_overview}
\end{figure}


\noindent 3D imaging is essential in several fields, from clinical applications with Magnetic Resonance Imaging (MRI) and Computed Tomography (CT) modalities \cite{corona2022mednerf, ying2019x2ct}, to virtual/augmented reality \cite{10.1145/3562939.3565632, park2022holographic}, self-driving vehicles \cite{li21durlar, Sautier_2022_CVPR}, and security \cite{wang20tip, wang20ct3d}. However, the diverse range of available imaging devices exhibit differences in cost, quality, and accessibility, which has led to an increased interest in leveraging 2D imaging for 3D acquisition. For instance, in hospitals, CT from biplanar X-rays could minimize patient exposure to a substantial dose of ionizing radiation \cite{COFFEY_2010}. Similarly, at airports, volumetric reconstruction from security baggage screening could be more effective at detecting prohibited items \cite{wang21segmentation, wang21materials}. The capability to seamlessly translate between these imaging domains, eliminating their limitations, is a high-impact application; however, this remains a difficult conditional generative modeling problem.

Generating accurate 3D representations from 2D images poses a significant challenge in the field of computer vision. Approaches must account for complex shape topologies, fine-grained textures, domain differences, and incomplete input information. Current methods aim to solve this under-constrained problem by relying on cycle consistency losses \cite{Zhou_2016_CVPR} for disentangling 3D shape properties like appearance and viewpoint from a single input image \cite{9320324}, improving fine-grained 3D mesh attributes \cite{2021_CVPR_3d_recon}, for 3D-aware face image generation by distilling StyleGAN2 latent space \cite{Shi_2021_CVPR} and recovering 3D shapes from videos \cite{NEURIPS2021_a11f9e53}. Despite these models being unpaired, they require separate convolutional neural network (CNN) designs for predicting each 3D property, and they cannot generate accurate 3D representations without additional 2D supervisory signals.

In this work, we present a unified approach that tackles 3D synthesis by reformulating it as a voxel-to-voxel prediction problem. We achieve this by conditioning an unconstrained transformer on 2D input views. While some recent methods have used transformers for multi-view 3D object reconstruction, they suffer from computational inefficiency and are sensitive to domain shifts and input view irregularities \cite{Wang_2021_ICCV, Reizenstein_2021_ICCV}.

We investigate 2D to 3D image translation of complex data that exhibit varying outer and internal topologies with different density properties and domains. To achieve this, we propose a novel two-stage translation approach that models the conditional probability of generating a 3D image given 2D input views with a discrete diffusion model. By applying this process in a highly compressed discrete latent space, our approach can extract high level features of complex objects and scale to high-resolution data without requiring paired datasets (Fig. \ref{fig:task_overview}), which is a desired feature for real-world applications. First, our approach learns information-rich discrete spaces of 2D and 3D distributions independently with two Vector-Quantized Variational Autoencoders (VQ-VAE), removing the need for alignment of both geometries.  Second, we use a diffusion model parameterized by an unconstrained transformer that allows bidirectional 2D global context when generating the 3D representation, improving feature learning and speeding up the sampling process.\\


To summarize, our main contributions are:
\begin{itemize}   
\item[--] We propose a novel and simple translation approach based on conditional diffusion using transformers, generating high-quality 3D samples conditional on two 2D images from a different domain.
\item[--] We show that diffusion in the information-rich latent code space not only allows for our model to scale easily to high-resolution, but also allows for translation of unaligned inputs---as our full-coverage attention on latent encodings permits any part of all 2D inputs to contribute to voxel predictions.
\item[--] The model is shown to give state-of-the-art density and coverage over competing methods such as Generative Adversarial Networks (GAN) and Neural Radiance Fields (NeRF) while offering true likelihood estimates.
\end{itemize}

\vspace{0.5cm}
\section{Prior Work} \label{s:relwork}
\subsection{Autoregressive Modeling}
\noindent Autoregressive models are a class of likelihood-based generative models that have been demonstrated to be potent density estimators, exhibiting greater training stability and generalization capabilities \cite{salimans2017pixelcnn,NIPS2016_b1301141} compared to implicit generative models such as GAN \cite{10.1145/3422622}. They break down the joint distribution of structured outputs into products of conditional distributions,
\begin{equation}
    p({\vec{c}}|\vec{Z})=\prod_{i=1}^L p_\theta(c_i | c_1,...,c_{i-1}; \vec{Z}).
\end{equation}
However, as their receptive field is limited to previously generated tokens, their representation ability is biased, and images do not conform to such sequential manner. Moreover, this also restricts them to relatively low dimensional data \cite{pmlr-v48-oord16,vandenOord_2016}.

\subsection{Vector-Quantized Representations}
\noindent The vector quantization (VQ) technique has been adopted by explicit generative models to alleviate issues including scaling, posterior collapse and blurred outputs by quantizing the latent representations to a fixed number $\{\vec{q}^1,\cdots,\vec{q}^K\}$ \cite{NIPS2017_vqvae, Han_2022_CVPR, NEURIPS2021_e46bc064}. Furthermore, VQ-based models have achieved sharper reconstructions than implicit generative models on continuous latent variables. Following their success in generative modeling, we make use of VQ representations, where a convolutional autoencoder extracts high-level features to an information-rich latent space. VQ image models \cite{NIPS2017_vqvae}, which compress images to a low dimensional discrete latent space and subsequently model this space with a powerful generative model, have recently been used for a variety of tasks. Chen et al., \cite{chen2022vector} address the domain gap issue in cross-domain analysis by introducing VQ into the image-to-image translation framework; however, their approach deals with the two data modalities having the same dimension and relies on spatial correlations. 

\begin{figure*}[!ht]
    \centering
    \includegraphics[width=0.98\linewidth]{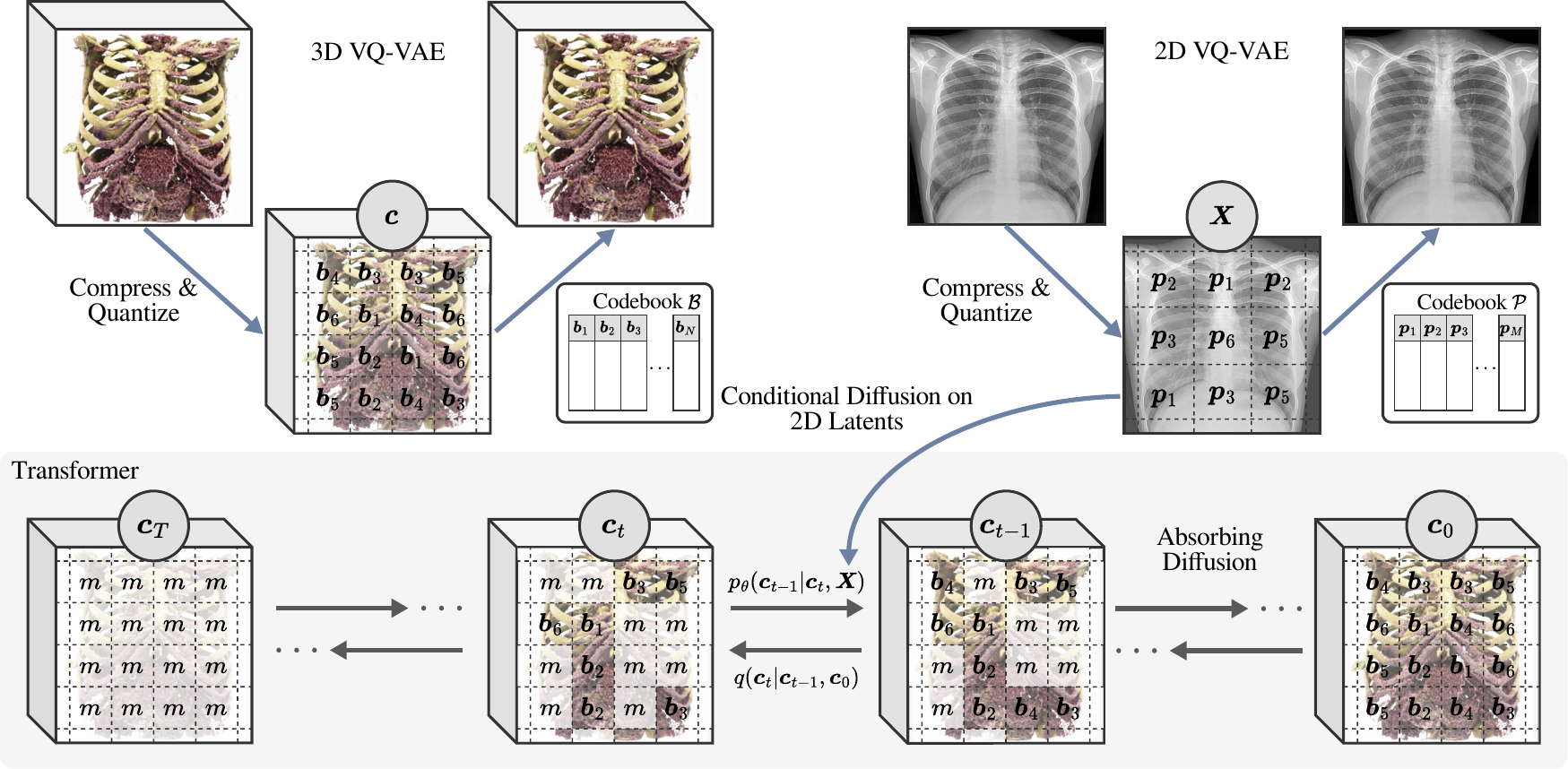}
    \caption{Our approach allows 2D to 3D translation from unaligned data by first compressing to an information rich Vector-Quantized discrete space, then modeling the conditional probability with a discrete diffusion model parameterized by an unconstrained transformer.}
    \label{fig:main-diagram}
\end{figure*}

\subsection{2D to 3D Image Translation}
\noindent Image-to-image translation methods which aim to reconstruct a 3D representation given a single or multiple 2D images, have achieved notable success within the field \cite{2021-3d-recon}. Generally, these architectures first extract 2D features from the input image into a latent vector, which can be fused with other information such as geometric priors, and lastly, the decoder generates the predicted 3D representation \cite{Tatarchenko_2019_CVPR, 2021-3d-deep-rev, 2021-3d-recon}. However, these techniques applied to natural images do not easily translate to real-application domains. Moreover, most conditional generative models have dealt with the input and output data having the same dimension, i.e. 2D to 2D or 3D to 3D.

\paragraph{Generative Adversarial Networks.} GAN \cite{10.1145/3422622} have emerged as representatives of implicit generative modeling due to their capability to generate realistic synthesized images. However, they often suffer from training instability \cite{arjovsky2017towards}. In our domain's context, we are comparing two GAN-based models. The first is X2CT-GAN \cite{ying2019x2ct}, which utilizes a 3D GAN with skip-connections, along with an additional CycleGAN for capturing style differences. The second model is CCX-rayNet \cite{isbi_2021_ccx}, which employs a class-conditioned module to reconstruct a CT from 2D X-ray images. Unlike CNN-based models that require aligned inputs (i.e. consistent resolution, orientation, and transformations) and often rely on skip-connections, our approach allows unaligned inputs via full-coverage attention on an information-rich discrete latent space.
\paragraph{Neural Radiance Fields.} The ability of NeRF to generate novel views of complex 3D scenes from a partial set of 2D images \cite{mildenhall2021nerf}, inspired MedNeRF \cite{corona2022mednerf} for rendering CT projections from a small set or single-view X-rays. While NeRF render 3D-informed images from 2D viewpoints, they differ fundamentally from our method. In particular, they are focused on object and scene representation conditional on coordinate information, rather than novel synthesis and generalizability \cite{gao2023nerf}. In contrast, our approach needs no prior data on camera positions or aligned inputs. It enables direct sampling based on input views while providing control over the generative process.

\subsection{Hybrid Generative Models}
\noindent A way to address issues in specific generative models such as long training or poor scaling is by combining two or more approaches \cite{bond2021deep}. For instance, transformers and their self-attention mechanism use in an encoder-decoder setup \cite{NIPS2017_3f5ee243} to improve both autoregressive models and other generative models due to their stable training and ability to learn long-distance dependencies. This is achieved using the attention scheme,
\begin{equation}
    \mathrm{Attention}(\boldsymbol{Q}, \boldsymbol{K}, \boldsymbol{V}) = \mathrm{softmax}\left(\frac{\boldsymbol{Q}\boldsymbol{K}^T}{\sqrt{d_k}}\right)\boldsymbol{V}.
\end{equation}
\noindent where the values $\boldsymbol{V}$ represent the encoded inputs, the keys $\boldsymbol{K}$ are used for indexing and queries $\boldsymbol{Q}$ to select specific values. Additionally positional information is passed into the function via fixed or trainable positional embeddings \cite{child2019generating}.

Of particular interest to this work is the two-stage process proposed by Bond-Taylor \textit{et al.} \cite{bond2022unleashing}, where the discrete latent space is modeled by a diffusion model parameterized by an unconstrained transformer that learns to unmask the data. This process allows faster sampling because multiple tokens can be predicted in parallel. Inspired by this work, we bridge the domain gap between 2D to 3D translation by combining the infinite receptive fields of attention for representing both data distributions and allowing learning complex topologies with the powerful feature-extraction ability and scalability of VQ-VAE. One of the advantages of our approach is that it does not require spatially aligned 2D and 3D samples, thus avoiding issues with both the geometric and domain misalignment between the two modalities.



\section{Method} \label{s:method}
\noindent In this work, we address the task of synthesizing complex 3D data portraying varying topologies and material properties (e.g. CT-like images) given a two 2D views taken from different angles (e.g. X-rays). In particular, we avoid dependencies on both the geometric and domain relationships between the 2D and 3D data which are significant sources of real-world data misalignment caused by object/device movement and characteristics. In this setting, learning a deterministic mapping between modalities is impractical due to a large number of possible input/output pairings in real-world data, meaning that outputs would be very blurry and unhelpful \cite{NEURIPS2018_801fd8c2} to human operators for instance. As such, we propose modeling the mapping between 2D and 3D data with a conditional likelihood-based generative model, allowing sampled 3D data to sit at arbitrary positions/rotations relative to the 2D data. 


Formally, given 3D data $\vec{I} \in \reals^{H \times W \times D}$, and multiple corresponding 2D views $\vec{X} = \{ \vec{x}_1, \dots, \vec{x}_n \}$, $\vec{x}_i \in \reals^{H \times W}$, we wish to learn the conditional distribution $p(\vec{I} | \vec{X})$. Training a transformer model directly on complex high-dimensional pixel data would be impractical, as the computational complexity would increase quadratically from the self-attention mechanism \cite{NIPS2017_3f5ee243, tay2021long}. To overcome this, we decompose the task into two stages to take advantage of the power and flexibility of transformers. Stage 1 separately learns to compress the 2D and 3D data to small but information-rich discrete latent spaces that accurately represent the data. In Stage 2, we model the conditional probability distribution in the learned compressed discrete space with a discrete diffusion model parameterized by a powerful unconstrained transformer that is spatially invariant over the 2D data. The overall process is illustrated in \cref{fig:main-diagram},  and train it alternatively with a patch-wise discriminator that uses a combination of spatial and style augmentations for both 2D and 3D images. More details in Appendix C.


\subsection{Stage 1: Unpaired Compression}
\noindent Separately for 2D and 3D data, we learn to compress a single data point $\vec{x}$ (each 2D view is also compressed separately) to a relatively small set of integer values $\vec{z}$. This is performed using a VQ-VAE \cite{NIPS2017_vqvae}, which achieves extremely high compression rates by utilising a codebook for information rich vectors, each of which is able to represent an image patch, while a neural decoder models how these codes interact.

The VQ-VAE approach is of particular interest to our work as it allow a 2-stage scheme where a compact and quantized latent space can be learned through a convolutional autoencoder with a large receptive field. In addition, variational autoencoders (VAE) \cite{vaes_2014} allow us to provision per-3D-image likelihood estimates, in contrast to other 2D-3D translation models based on GAN, which cannot directly provide a probability interpretation.



In particular, a convolutional encoder downsamples data $\vec{x}$ to a smaller spatial resolution, $E(\vec{x}) = \{\vec{e}_1, \vec{e}_2, ..., \vec{e}_L\} \in \mathbb{R}^{L \times D}$. Each continuous encoding $\vec{e}_i$ is subsequently quantized by mapping to the closest element in the codebook of vectors $\mathcal{B} \in \mathbb{R}^{K \times D}$, where $K$ is the number of discrete codes in the codebook, and $D$ is the dimension of each code,
\begin{equation}\label{eqn:quantisation}
    \vec{z}_q = \{\vec{q}_1, \vec{q}_2, ..., \vec{q}_L\} \text{  , where  } \vec{q}_i = \underset{\vec{b}_{j} \in \mathcal{B}}{\operatorname{min}}\|\vec{e}_i - \vec{b}_j\|,
\end{equation}
with the straight-through gradient estimator \cite{bengio2013estimating} used to approximate the gradients through this non-differentiable process. The discrete latents are then decompressed with a convolutional decoder $\hat{\vec{x}}= G(\vec{z}_q)$. The model is trained end-to-end by minimizing the loss,
\begin{equation}\label{eqn:vqvae-loss}
    \loss_\text{VQ} = \loss_\text{rec}(\vec{x}, \hat{\vec{x}}) + \| \text{sg}[E(\vec{x})] - \vec{z}_q \|_2^2 + \beta \| \text{sg}[\vec{z}_q] - E(\vec{x}) \|_2^2.
\end{equation}
which balances reconstruction quality $\loss_\text{rec}$ against quantization terms that encourage the encoding $E(\vec{x})$ to match the closest codes in the codebook.

Our approach eliminates the need for spatially aligned 2D and 3D images, which in practice are difficult to obtain accurately. In addition, the flexibility of our model allows the use of an arbitrary number of 2D input views without the requirement of camera priors or changing the network architecture as this can be effectively achieved by using smaller or larger latent code sizes. An ablation of these aspects can be found in the Appendix A.

%

\subsection{Stage 2: Conditional Discrete Diffusion} 

\noindent To translate data from 2D to 3D we model the conditional probability distribution of 3D data given a few 2D views using a discrete diffusion model in the learned vector-quantized space, $p(\vec{c}|\vec{X})$, where $\vec{c}$ represents the VQ codes of the 3D data and $\vec{X} = \{ \vec{x}_1, \dots, \vec{x}_n \}$ represents the set of VQ codes of the all the 2D views. Specifically, we use the discrete absorbing diffusion process \cite{austin2021structured,bond2022unleashing}, which is more suitable for this task due to its bidirectional nature, allowing them to outperform autoregressive approaches while also being able to scale to higher dimensional spaces, faster sampling, and being substantially less prone to overfitting \cite{chang2023design} which is crucial when training on small datasets (as is generally the case for real-world datasets). More detail on this aspect is presented in the Appendix A (Fig. 4).

In this case, the discrete 3D latents are gradually masked out using randomized orders over a number of time steps $T$ such that at time step $t$, $\vec{c}^t$ is defined as the discrete 3D latent $\vec{c}$ with each token masked out with probability $\frac{t}{T}$. Formally, this masking procedure is defined by a Markov chain,
\begin{equation}
    q(\vec{c}^{1:T}|\vec{c}^0) = \prod_{t=1}^T q(\vec{c}^t| \vec{c}^{t-1}).
\end{equation}
The posterior is defined as $q(\vec{c}_t|\vec{c}_{t-1})= \text{cat}(\vec{c}_t;\vec{p}=\vec{c}_{t-1} \vec{Q}_t)$ for one-hot $\vec{c}_{t-1}$ where $\vec{Q}_t = (1-\beta_t)I + \beta_t \mathbbm{1}e_m^T$ is a matrix denoting the probabilities of moving to each successive state, $e_m$ is a vector with a one on mask states $m$ and zeros elsewhere, and $\beta_t = \frac{1}{T-t+1}$.

The reverse of this diffusion process is another Markov chain that gradually unmasks the latents
\begin{equation}
    p_\theta(\vec{c}^{0:T}|\vec{Z}) = \prod_{t=1}^T p_\theta(\vec{c}^{t-1}|\vec{c}^t,\vec{X}).
\end{equation}
This can be approximated by training an unconstrained transformer to predict the original latents from the masked ones, optimizing the Evidence Lower Bound (ELBO),
\begin{equation}
     \E_{q(\vec{c}^0,\vec{X})} \sum_{t=1}^T \gamma \E_{q(\vec{c}^t|\vec{c}^0)} \Big[ \sum_{[\vec{c}^t]_i=m}\hspace*{-0.3em}\log p_\theta([\vec{c}^0]_i|\vec{c}^t,\vec{X}) \Big],
\end{equation}
where $\gamma=\frac{T-t+1}{T}$ is a reweighting term used to improve convergence \cite{bond2022unleashing}.
\paragraph{Discrete vs. Continuous Latents} Discrete representations are important for our approach because applying self-attention on a compact, information-rich space is more efficient than attending over a larger, continuous space due to limitations in sequence length \cite{tay2021long} coupled with less effective integration of information \cite{mao22:discrete}. Moreover, compressed representations have been shown to improve generalization without relying on complex hierarchical architectures \cite{Wang_2023_CVPR}.
\vspace{-0.4cm}
\paragraph{Full-coverage Attention}
To leverage the compressed latent space from the diffusion model, we employ an unconstrained transformer to parameterize the denoising function. By flattening and concatenating 2D latents, we integrate discrete representations into a 1D manifold, aided by a trainable positional embedding. In contrast to CNN reliance on local alignment and/or deep architectures to increase their limited receptive fields, our method provides global context through information-rich discrete representations, allowing all parts of the 2D inputs to influence voxel predictions \cite{mao22:discrete}.
\vspace{-0.4cm}
\paragraph{Domain Invariance} Our approach is robust to differences in the distributions of 2D and 3D data domains as we model distributions over discrete latent representations separately rather than low-level voxels in a joint manner. In contrast, a unified framework such as GAN cannot achieve this unless they explicitly define a domain-invariant joint latent distribution. Domain invariance is of particular interest when dealing with real-world scenarios where data exhibits complex variations, such as different camera perspectives, lighting conditions, or imaging modalities. Examples of this are presented in the Appendix A (Fig. 4).
\vspace{-0.2cm}
\paragraph{Likelihood Estimation} Due to the fact that the VQ-VAE decoder is trained with MAP-inference, within this framework we are able to estimate the conditional log likelihood $\log p(\vec{I}|\vec{X}) \approx \log p(\vec{I}|\vec{c})p(\vec{c}|\vec{X})$ \cite{NIPS2017_vqvae}, where $p(\vec{I}|\vec{c})$ is the 3D VQ-VAE decoder, and $p(\vec{c}|\vec{X})$ is the conditional discrete diffusion model. Subsequently, our model provisions per-3D-image likelihood estimates, which provide a distance measurement from the true distribution of the data.



\begin{figure*}
    \centering
    \includegraphics[width=\linewidth]{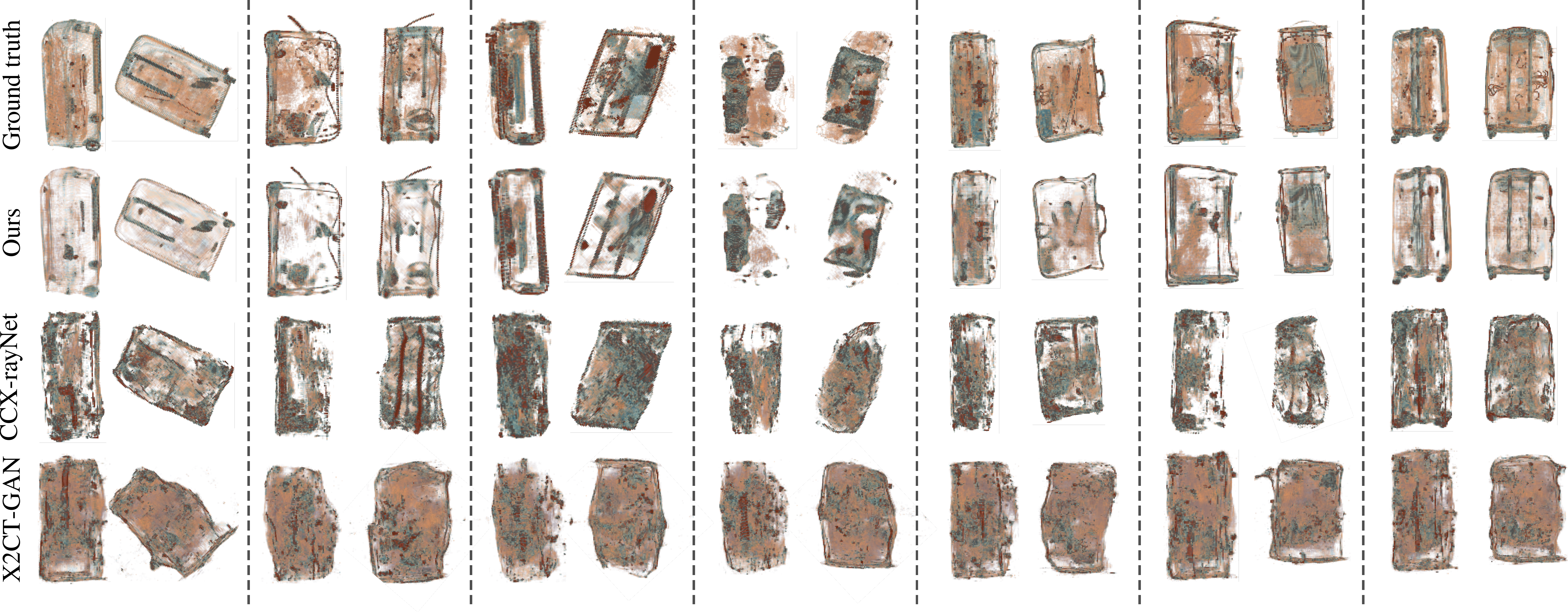}
    \vspace{-0.6cm}
    \caption{Comparison of 3D CT model samples trained on the baggage screening dataset, showing the ground truth, our method, CCX-rayNet \cite{isbi_2021_ccx} and X2CT-GAN \cite{ying2019x2ct}.}
    \label{fig:bags_qual_comp}
\end{figure*}

\section{Experiments} \label{s:experiments}

\noindent This section evaluates the ability of our unconstrained transformer to perform 2D to 3D translation using discrete VQ representations against the SOTA models on this task: X2CT-GAN \cite{ying2019x2ct} and CCX-rayNet \cite{isbi_2021_ccx} on two datasets complex volumetric imagery from the publicly available chest CT scans, LIDC-IDRI \cite{armato2011:lidc}, and security baggage screening. Additionally, we provide a 2D evaluation compared to the MedNeRF \cite{corona2022mednerf} model in terms of MIPs in Sec. \ref{ss:2d_eval}. We evaluate in terms of negative log-likelihood (NLL), Density and Coverage, and distortion metrics, including SSIM, PSNR, MSE and MAE. The best values are in bold, and the second-best values are underlined.

\begin{table*}
\begin{center}
\fontsize{9pt}{9pt}\selectfont
    \begin{tabular}{lccc|cccc}
    \toprule[1.5pt]
    \multicolumn{8}{c}{Baggage Security Screening dataset}
     \\ \midrule
              Method    & $\downarrow$ NLL 
           & $\uparrow$   Density    & $\uparrow$ Coverage
           & $\uparrow$   SSIM & $\uparrow$   PSNR
           & $\downarrow$ MSE  & $\downarrow$ MAE
    	\\    \midrule 
    X2CT-GAN   & $\text{N/A}$   
               & 0.95   & 0.80
               & 0.655    & 34.68
               & 0.0014   & 0.0129
             \\
    CCX-rayNet & $\text{N/A}$   
               & 1.28                 & \underline{0.89}
               & \underline{0.886}    & \underline{35.45}
               & \underline{0.0012}   & \underline{0.0069}
             \\ 
    Ours       & \textbf{0.007}    
               & \textbf{2.01}      & \textbf{0.97}
               & \textbf{0.899}     & \textbf{39.45}
               & \textbf{0.0007}    & \textbf{0.0049}
             \\
    \bottomrule[1.5pt]
    \end{tabular}
\end{center}
\caption {\textbf{3D fidelity and distortion metrics on the Baggage security dataset.} We compare our method and SOTA models in terms of fidelity and diversity (density and coverage). Additionally we evaluate the quality of generated voxel grids using distortion metrics (SSIM, PSNR, MSE and MAE).}
\label{tab:metrics_comp_bags}
\end{table*}

\subsection{Conditional 3D Modeling on 2D Views}\label{ss:eval_task}
\noindent 2D to 3D translation is performed by our unconstrained transformer trained on the discrete latents encodings computed by the two VQ-VAE. At inference, the 3D latents are gradually predicted using the conditioning information from two 2D inputs in the form of $16\times16$ codes, which are concatenated at the start.
\paragraph{Fidelity Metrics.} We aim to measure the quality as well as the overlap between the manifold of generated samples and the manifold of real data. One of our primary motivations for using likelihood-based generative models is that Negative Log-Likelihood (NLL) allows us to monitor over-fitting effectively. This is a potential issue when training models on real application datasets which are generally much smaller than natural image datasets. In addition, we report values based on Density-Coverage \cite{naeem20_dens_cov}, which independently assesses the fidelity and variability of a model. In these metrics, images are first projected into an embedding space, and a scoring function estimates the manifold density in the neighborhood of each embedded data point. Generative models trained with natural images mostly rely on the features from an ImageNet pretrained model for evaluation. Since our target data is distinct from ImageNet samples, to compute these metrics, we use the features of a randomly initialized convolutional encoder architecture, as proposed in the work of \cite{naeem20_dens_cov}. Specifically, we obtained embeddings of dimension 2,197 of the real validation data points and samples from our model and competing models.
\vspace{-0.3cm}
\paragraph{Distortion Metrics.} We emphasize that commonly used distortion metrics such as Peak Signal-to-Noise Ratio (PSNR) and Structural Similarity Index (SSIM) correlate poorly with human visual perception, as they ignore global structure and only focus on signal fidelity \cite{2021_img_qual}. A model that simply optimizes pixel-level distortions (e.g. small adjustments in brightness, saturation etc.) can obtain a high score on these metrics. However, we also report values on SSIM, PSNR, Mean Squared Error (MSE) and Mean Absolute Error (MAE) for the research community as they are widely used for comparison \cite{ying2019x2ct, isbi_2021_ccx}.

\subsubsection{Modeling Objects in Baggage Screening}\label{ss:bags_eval}
We evaluate our approach on a set of \textit{``stream of commerce''} CT volumes from an aviation security context. The total dataset consists of 5,964 CT bag volumes. We resample the bag volume to $1\times1\times1$ mm$^3$, and subsequently, we resize it to a $256\times256\times256$ mm$^3$ cubic area. This dataset comprises authentic bags and suitcases, depicting both the common and the forbidden items during air travel. We generated a 2D subset of eight views, each taken at 45-degree azimuth intervals per bag. For our model, we set $|\vec{b}_i|=256$, a codebook of 4,096 and train our transformer to predict sequences of length $16\times16\times16$.

\begin{figure*}[t]
    \centering
    \includegraphics[width=\linewidth]{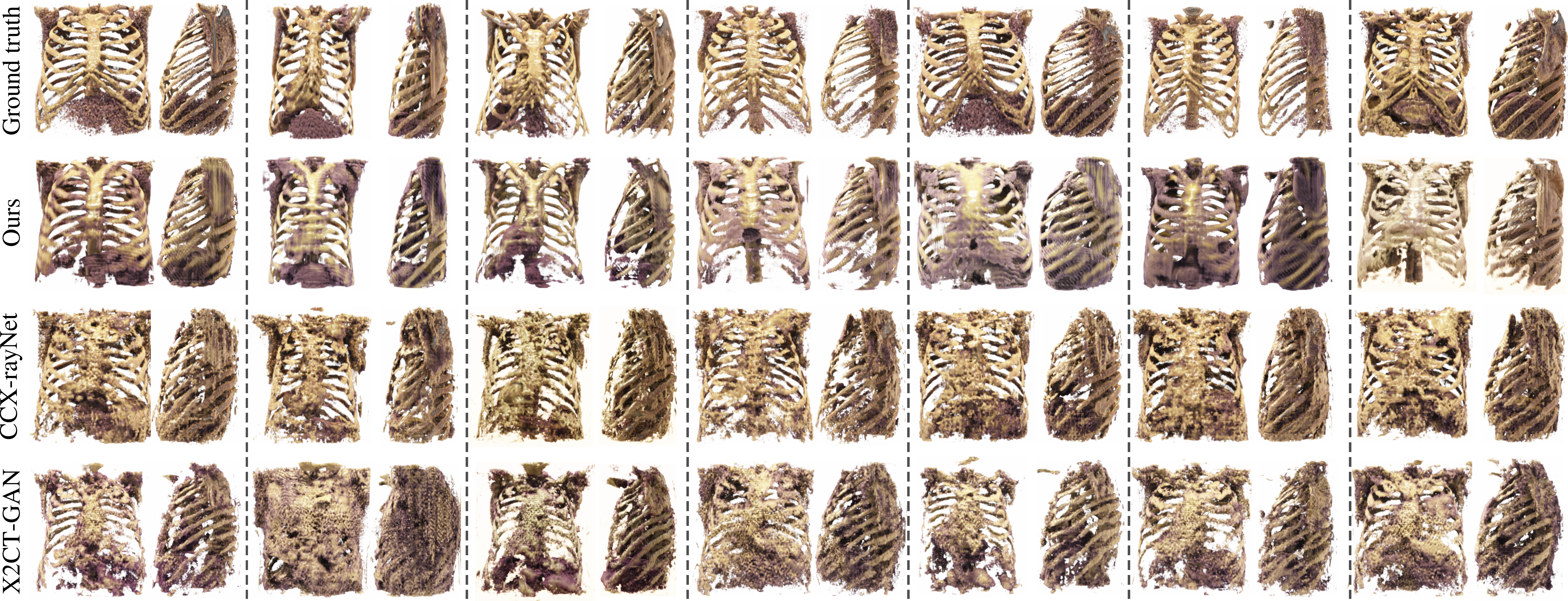}
    \caption{Comparison of 3D CT model samples trained on the medical chest (LIDC-IDRI dataset \cite{armato2011:lidc}) from different anatomical planes (coronal and sagittal), showing the ground truth, our method, CCX-rayNet \cite{isbi_2021_ccx} and X2CT-GAN \cite{ying2019x2ct}.}
    \label{fig:chest_qual_comp}
\end{figure*}

\begin{table*}
\begin{center}
\fontsize{9pt}{9pt}\selectfont
    \begin{tabular}{lccc|cccc}
    \toprule[1.5pt]
    \multicolumn{8}{c}{LIDC-IDRI (chest) dataset}
            \\ \midrule
     Method    & $\downarrow$ NLL 
               & $\uparrow$   Density    & $\uparrow$ Coverage
               & $\uparrow$   SSIM & $\uparrow$   PSNR
               & $\downarrow$ MSE  & $\downarrow$ MAE
    	\\    \midrule 
    X2CT-GAN   & $\text{N/A}$     
               & 0.87   & 0.88
               & 0.321   & 19.68
               & \underline{0.045}   & 0.151
             \\
    CCX-rayNet & $\text{N/A}$      
               & \underline{1.41}   & \textbf{0.98}
               & \underline{0.386}  & \underline{22.66}
               & \textbf{0.032}  & \underline{0.108}
             \\
    Ours       &  \textbf{0.10}    
               & \textbf{1.42}     & \underline{0.97}
               & \textbf{0.436}    & \textbf{25.05}
               & 0.048    & \textbf{0.013}
             \\   
    \bottomrule[1.5pt]
    \end{tabular}
\end{center}
\caption {\textbf{3D fidelity and distortion metrics on LIDC-IDRI (chest) dataset.} We followed the experimental protocol X2CT-GAN \cite{ying2019x2ct} and CCX-rayNet \cite{isbi_2021_ccx} with a wider Hounsfield unit range of -1,000 HU to +1,000 HU. We show additional analysis on out-of-distribution inputs in the Appendix A (Fig. 4).}
\label{tab:metrics_comp_chest}
\vspace{-0.3cm}
\end{table*}
\paragraph{Results.} Table \ref{tab:metrics_comp_bags} reports results for conditional 3D modeling on two 2D input views comparing samples from our model and reconstructions from the competing approaches \cite{ying2019x2ct} and \cite{isbi_2021_ccx} for the baggage dataset. This dataset depicts varying topologies as there exists a wide variety of baggage, such as suitcases, and backpacks, in different sizes, styles, and materials. Moreover, the items/objects within passenger bags can be unpredictable and their arrangement and level of compactness could complicate their identification. Despite the challenging complexity of this dataset, we find that our approach offers superior performance, with a very large margin in terms of density ($+111\%$ improvement over X2CT-GAN \cite{ying2019x2ct}, and $+53\%$ over CCX-rayNet \cite{isbi_2021_ccx}) and coverage ($+21\%$ over X2CT-GAN and $+8\%$ over CCX-rayNet). In terms of distortion metrics, our samples are of higher quality, with a significant improvement in PSNR, MSE and MAE compared to the competing models \cite{ying2019x2ct,isbi_2021_ccx}.

In Fig. \ref{fig:bags_qual_comp} we present a set of 3D generated samples from our approach and 3D reconstructions of the other models for the baggage dataset. It can be observed that the identity of each of the bags and suitcases is accurately modeled by our model, while CCX-rayNet \cite{isbi_2021_ccx} is limited at predicting the shape boundaries, and X2CT-GAN \cite{ying2019x2ct} shows a lack of diversity. It is worth noting that airport security officials focus on denser objects as these are more likely to be prohibited compared to less dense items such as clothes. In this context, our model is better able to model latent representations of denser structures, as indicated by darker colors, which aligns with real-world scenarios. In contrast, the other models depict blurring for both dense and soft structures, making it challenging to distinguish them. 

\subsubsection{Modeling Chest Anatomical Structures}\label{ss:chest_eval}
We conduct a set of experiments using the publicly available dataset of chest CT scans, LIDC-IDRI \cite{armato2011:lidc}. We generated the corresponding X-ray projections using digitally reconstructed radiograph technology (DRR). For our model we set $|\vec{b}_i|=256$, a codebook of 1,094 and train our transformer to predict sequences of length $8\times8\times8$.
\vspace{-0.4cm}
\paragraph{Results.} As presented in Table \ref{tab:metrics_comp_chest}, our method exhibits superior performance across almost all categories for density and coverage and surpasses the other models in all distortion metrics, achieving a $12\%$ improvement in SSIM, $11\%$ in PSNR, $41\%$ in MSE, and $28\%$ for MSE over the most competitive model, CCX-rayNet \cite{isbi_2021_ccx}. Furthermore, in comparison to X2CT-GAN \cite{ying2019x2ct}, our model demonstrates an even greater degree of improvement performance.

\begin{figure}[t]
\centering
\includegraphics[width=\columnwidth]{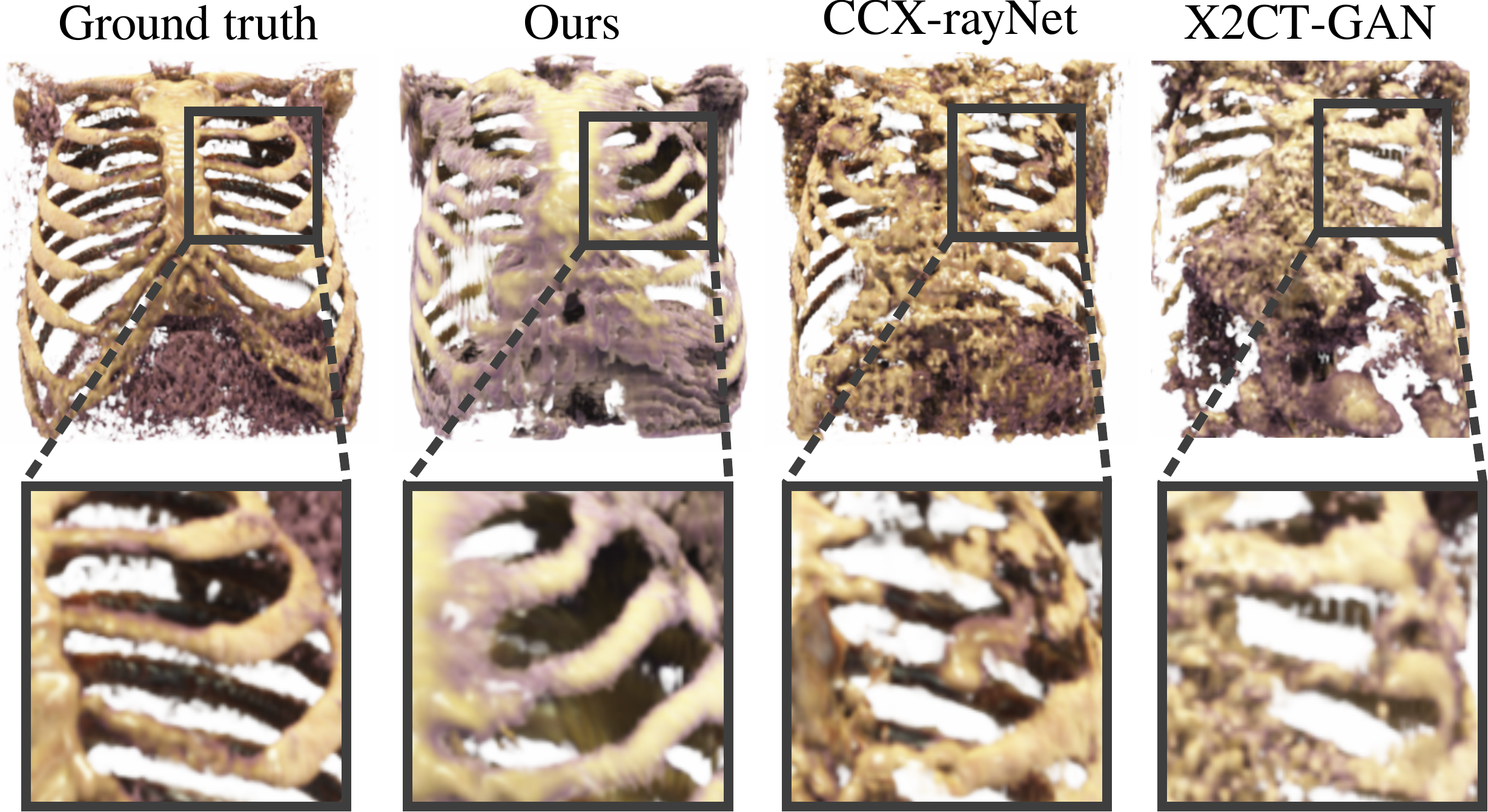}
\caption{Local-level quality comparison of chest CT samples.}
\label{fig:chest_zoom}
\end{figure}

In Fig. \ref{fig:chest_qual_comp} we present a set of 3D generated samples from our approach and 3D reconstructions of the other models for the LIDC (chest) dataset. Although the sizes of both datasets are relatively small (compared to natural image datasets), our model displays no signs of overfitting and can produce diverse shapes that closely align with the ground truth. In particular, our approach achieves a greater degree of precision in modeling and disentangling soft tissue and bone structures. This is evidenced by the distinctive colors employed in the renderings, with a purple\textit{ish} tone denoting soft tissue and a beige color representing bone. By contrast, the other methods produce blurred images where anatomical structures overlap, and as a result, densities cannot be accurately modeled since they are all merged together. For instance, samples from our approach are less noisy, while the other models are only able to reasonably predict some parts of the volume and result in spatial blur/uncertainty in others (Fig. \ref{fig:chest_zoom}).

\subsubsection{Analysis on Maximum Intensity Projections}\label{ss:2d_eval}
We perform an additional evaluation in terms of Maximum Intensity Projections (MIP) in 2D, as they provide a representation of the highest intensity values along the spatial dimensions, accentuating low-level structures, which might be difficult to detect on individual slices of the 3D data. MIP are frequently used in real-world settings for clinical diagnosis, planning, object identification, or to simply enhance visibility. For our case, this could provide an additional interpretation of both quantitative and qualitative results.

In addition to X2CT-GAN \cite{ying2019x2ct}, and CCX-rayNet \cite{isbi_2021_ccx}, we also compare our model with the NeRF-based method MedNeRF \cite{corona2022mednerf}, which has been designed to handle complex images such as X-rays. For this, we train MedNeRF \cite{corona2022mednerf} on both datasets and test them to render 3D-aware CT projections. For the other models, we obtained the MIP of their spatial dimensions and calculated density and coverage, SSIM and PSNR. These results can be found in Table \ref{tab:2d_comp}. Qualitative results of this comparison are presented in the Appendix B.

\subsection{Implementation Details}
\noindent  The 2D and 3D VQ-VAE from the first stage were trained simultaneously on independent NVIDIA A100 cards with a batch size of 32 and 8, respectively. For the VQ-AE we use the framework proposed in the work of \cite{esser2021taming}, which substantially improves compression rates by learning a more information-rich codebook, while still permitting likelihood estimates. The 2D model takes approximately half the time of the 3D model to complete 100,000 iterations. Our unconstrained transformer from the second stage takes less than 5 hours to also complete 100,000 iterations; it effectively models 3D tokens very quickly. We use weights from 500\textit{k} iterations for our reconstructions (Table \ref{tab:performance}). The transformer can easily fit into memory on a GPU with 12GB of VRAM with a batch size of 20 while the VQ-VAEs can also fit in such a GPU. In practice, this requires the use of small batch sizes making training in reasonable times less practical.
\begin{table}[t]
\begin{center}
\fontsize{9pt}{9pt}\selectfont
    \begin{tabular}{lccccc}
    \toprule[1.5pt]
    \multicolumn{5}{c}{(a) Baggage Security Screening dataset}
            \\ \midrule
     Method    & $\uparrow$ D  & $\uparrow$ C
               & $\uparrow$ SSIM & $\uparrow$ PSNR
    	\\  \midrule
    X2CT-GAN  \cite{ying2019x2ct} 
               & 0.51           & 0.68   
               & 0.65         & 34.49
             \\ 
    CCX-rayNet \cite{isbi_2021_ccx}
               & \underline{0.96}           & \underline{0.95}   
               & \underline{0.88}         & \underline{35.49}
             \\ 
    MedNeRF \cite{corona2022mednerf}
               & 0.91          & 0.46   
               & 0.79        & 25.11
             \\ 
    Ours  
               & \textbf{1.84} & \textbf{0.99}   
               & \textbf{0.91}         & \textbf{39.43}
                 \\ \toprule[1.5pt]

    \multicolumn{5}{c}{(b) LIDC-IDRI (Chest) dataset} \\ \midrule

    X2CT-GAN \cite{ying2019x2ct}
               & \underline{0.96}    & 0.85   
               & 0.32              & 21.38
             \\ 
    CCX-rayNet \cite{isbi_2021_ccx}
               & 0.76                & \underline{0.87}   
               & \underline{0.40}  & 24.20
             \\ 
    MedNeRF \cite{corona2022mednerf}   
               & 0.90           & 0.80   
               & 0.38         & \textbf{27.02}
             \\ 
    Ours     
               & \textbf{1.17}           & \textbf{0.91}
               & \textbf{0.42}         & \underline{25.25}
    
   \\ \bottomrule[1.5pt]
    \end{tabular}
\end{center}
\caption{\textbf{2D (Maximum Intensity Projections) fidelity and distortion metrics on both datasets.} We compare MedNeRF \cite{corona2022mednerf} in addition to SOTA models and our approach.}
\label{tab:2d_comp}
\end{table}


\begin{table}[t]
\begin{center}
\fontsize{9pt}{9pt}\selectfont
    \begin{tabular}{lcc}
    \toprule[1.5pt]
     Component  & Training time (hs) & Inference time (s)
    	\\    \midrule 
    2D VQ-VAE   & $\sim$ 23 & $\sim$ 0.01
             \\
    3D VQ-VAE   & $\sim$ 56 & $\sim$ 0.27
             \\
    Transformer & $\sim$ 4.5 & $\lesssim$ 10.9
   \\ \bottomrule[1.5pt]
    \end{tabular}
\end{center}
\caption{Training times (each for 100k iterations) and inference times for our approach, all performed on single NVIDIA A100 GPUs. The VQ-VAEs are independent and can be trained simultaneously on different GPUs.}
\label{tab:performance}
\end{table}





\vspace{5pt}
\section{Conclusion} \label{s:conclusion}
\noindent We propose a novel 2D to 3D translation approach based on conditional diffusion using transformers. We find that compressing each domain independently offers several key advantages. The discrete compressed space allows for both fast and high-resolution image synthesis, where the 2D and 3D compression networks can be trained and verified independently without requiring aligned datasets. In particular, full-coverage attention over the complete information-rich 2D codebooks from a few views significantly improves the synthesis of new 3D images, where any part of all 2D inputs can contribute to the voxel predictions.

The proposed approach is surprisingly simple and intuitive in practice as diffusion models are shown to have excellent mode coverage giving diverse samples \cite{chang2023design}. 
In the future, we would like to consider scaling our approach with larger models trained on more diverse datasets, provide an in-depth study on hallucinated outputs, and see how well it can generalize between very different imaging modalities.

\vspace{1.8cm}
\paragraph{Acknowledgments} This work was supported by CONAHCyT, and by the EPSRC NortHFutures project (ref: EP/X031012/1).

\newpage
{\small
\bibliographystyle{ieee_fullname}
\bibliography{bib/egbib}
}
\clearpage


\section*{Supplementary material}

# Unaligned 2D to 3D Translation with Conditional Vector-Quantized Code Diffusion using Transformers

Abril Corona-Figueroa[1]   Sam Bond-Taylor[1]   Neelanjan Bhowmik[1]
Yona Falinie A. Gaus[1]   Toby P. Breckon[1,2]   Hubert P. H. Shum[1]   Chris G. Willcocks[1]
Department of {[1]Computer Science | [2]Engineering}, Durham University, Durham, UK

The supplementary material for our work is structured as follows: First, Sec. A provides results for our ablation experiments including latent code sizes, codebook length and number of input views. Next, in Sec. B we present additional qualitative results showcasing: MedNeRF comparison, using out-of-distribution inputs, training on binary data, data augmentation details, and sampling using the autoregressive method. Finally, Sec. D includes code snippets for both the conditional diffusion process and for sampling from our model.

## A. Ablations

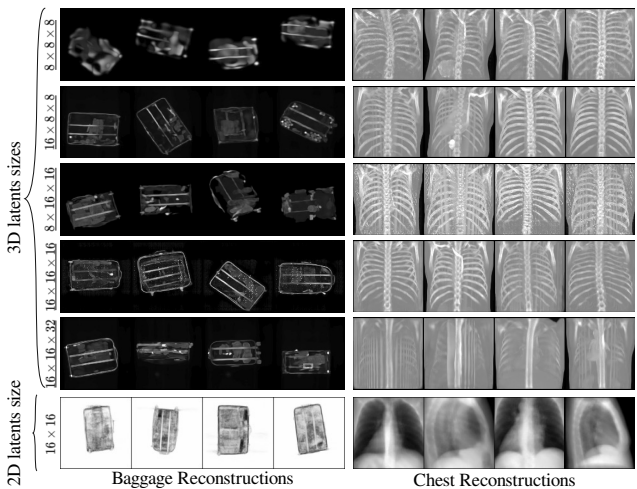

Figure 1. **Comparison of reconstruction quality using different latent code sizes on the two datasets.** 3D latents (ablation) correspond to the 3D VQ-VAE and 2D latents (fixed) to the 2D VQ-VAE. The chest dataset shows faithful reconstruction even with a small size, whereas the baggage security screening dataset requires a more complex latent representation. Larger latent code sizes could allow the model to learn more complex structures (baggage case) or mislead to learn unhelpful features (chest case).

|  | ↓ NLL | |
|---|---|---|
| Latent code sizes | Chest | Baggage |
| $8 \times 8 \times 8$ | **0.031** | 0.0087 |
| $16 \times 8 \times 8$ | 0.032 | 0.0056 |
| $8 \times 16 \times 16$ | 0.029 | 0.0055 |
| $16 \times 16 \times 16$ | 0.032 | **0.0044** |
| $16 \times 16 \times 32$ | 0.067 | 0.0056 |

Table 1. Quantitative evaluation based on validation Negative Log-Likelihood (NLL) using different latent code sizes for the discrete representations learned by the two VQ-VAE in Stage 1 of our approach.

| Codebook, $\mathrm{argmin}(\boldsymbol{x}, \mathcal{B})$ | | | |
|---|---|---|---|
| Length | Chest ($10^{-3}$) | Length | Baggage ($10^{-5}$) |
| 64 | 4.01 | 1024 | 5.8 |
| 128 | 0.85 | 2048 | 4.2 |
| **512** | 0.65 | **4096** | 1.7 |

Table 2. Quantitative evaluation based on the codebook loss using different codebook lengths for the discrete representations learned by the two VQ-VAE in Stage 1 of our approach. Our approach allows learning complex data distributions (e.g. baggage security dataset) by increasing the length. For our main experiments we used a length of 512 and 4,096 for the chest and baggage dataset, respectively.

Figure 2. **Ablation using different number of input 2D views for conditional 3D modeling on the LIDC-IDRI (chest) dataset.** While increasing the number of inputs views from 2 to 4 brings additional performance, further increments don't necessarily result in linear returns. For our main experiments we used only 2 inputs views for both datasets.

(input views: GT, 2, 4, 8; ↓ re-weighted ELBO: 1.163, 0.930, 0.929)

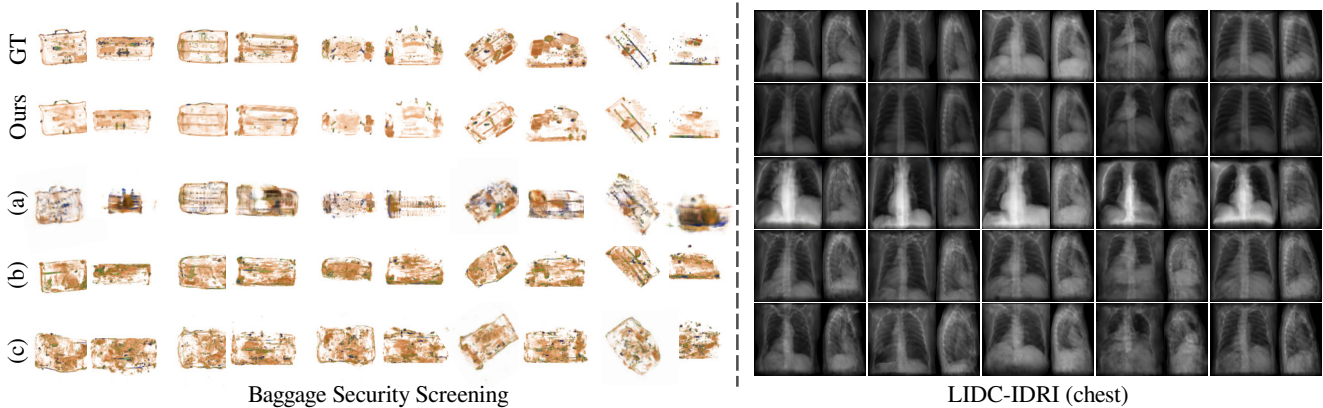

Figure 3. Comparison of Biplanar Maximum Intensity Projections (MIP) on the baggage security screening dataset and on LIDC-IDRI (chest) dataset. (a) denotes the MedNeRF model, (b) CCX-rayNet and (c) X2CT-GAN.

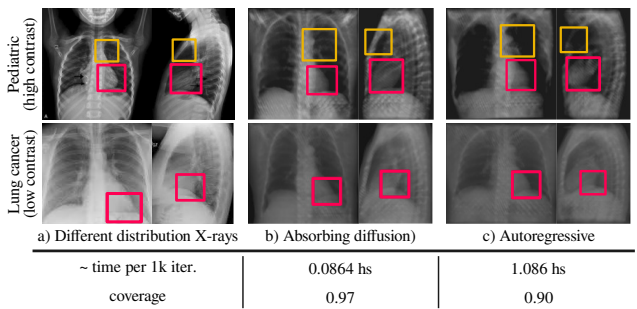

Figure 4. **Absorbing diffusion vs autoreggressive samples on out-of-distribution input X-rays.** We test our model on different distributions including pediatric and lung cancer X-rays with different radiographic contrast (high and low contrast).

## B. Additional Results

**2D Comparison** Fig. 3 contains qualitative results for both datasets of the 2D evaluation that also includes the MedNeRF method, which renders 3D-aware CT projections. This evaluation consists of comparing Maximum Intensity Projections from the spatial dimensions. Our method allows faithful samples compared to competing methods. Specifically, our samples better highlight denser objects which are of particular interest in the detection of prohibited items.

**Diffusion comparison** We test our model robustness on out-of-distribution input X-rays and generate samples comparing absorbing diffusion and the autoregressive method using the same transformer architecture. Despite domain differences, our model is able to generate accurate samples without requiring any kind of domain supervision, suggesting that our learned discrete representations achieve effective compression to remain invariant to low-level features like contrast while encoding essential structures such as bone and soft tissue (Fig. 4).

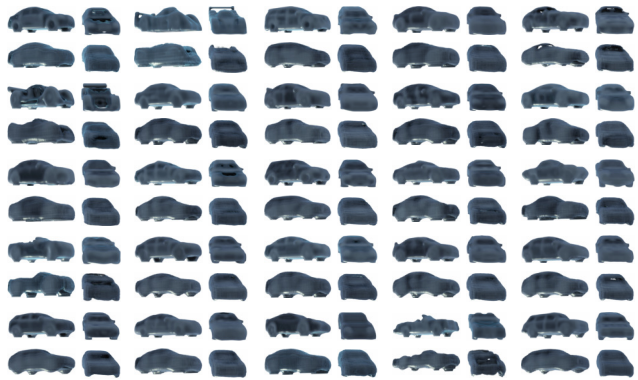

Figure 5. **Visualization of additional results on ShapeNet [3].** Training on binary volumetric data in contrast to continuous intensities isn't trivial as it might lead to instabilities in architectures like simple CNN-based GAN. Voxels with values other than 0 or 1 are flagged as fake by the discriminator, hindering preventing continuous optimization [1]. However, our model can effectively learn this type of data as our approach doesn't (necessarily) rely on adversarial training.

## C. Controlled Data Augmentations

To avoid overfitting, we extended the stochastic discriminator augmentation framework from StyleGAN2-ADA [2] to handle 3D data. This solution involves augmenting both real and generated data by the VQ-VAE using a set of style and spatial augmentations with a probability $p < 1$. Unlike other data augmentation strategies, non-invertible augmentations can be incorporated with an adaptive $p$-value based on an overfitting heuristic. We found that this prevents the discriminator becoming more confident, and thus both real and fake predictions take more time to diverge. As a result, both VQ-VAE are able to learn richer representations while delaying the drop in its the validation accuracy. Note that our approach doesn't rely on adversarial training, thus, the incorporation of a discriminator is optional.

## D. Example Code

We include python-like code for training a conditional absorbing diffusion process Fig. 6a, and sampling from our 2D to 3D translation model Fig. 6b. The use of a Transformer allows the learned distribution to be easily conditioned on arbitrary input shapes, by simply concatenating the conditioning signal with the noisy data. The linear masking schedule allows sampling with smaller numbers of steps, to speed the process up, by simply passing in a smaller value for $T$.

```python
def diffusion_training_loss(c_0, Z, T, mask_id):
    c_t, b = c_0.clone(), c_0.size(0)
    # Randomly sample diffusion time steps
    t = torch.randint(1, T+1, (b,))
    # Randomly mask tokens with probability t/T
    mask = torch.rand_like(c_0) < (t / T)
    c_t[mask] = mask_id
    # Calculate p(c_0 | c_t, Z)
    logits = Transformer(torch.cat(Z, c_t))
    # Calculate reweighted ELBO loss
    loss = cross_entropy(logits, c_0) * (T-t+1)/T
    return loss
```

(a) Python-like code snippet for training a conditional Absorbing Diffusion model.

```python
def sample(imgs_2d, T, mask_id, latent_size):
    b = imgs_2d.size(0)
    # Compress 2D images with 2D VQ-Encoder
    Z = vqae_2d.encoder(imgs_2d)
    # Initialise 3D latents with all masks
    c_t = torch.full((b, latent_size), mask_id)
    # Track which latents have been unmasked
    unmasked = torch.zeros_like(c_t, dtype=bool)

    # Loop over sampling steps
    for t in reversed(list(range(1, T+1))):
        # Randomly choose where to unmask
        changes = torch.rand(c_t.shape) < 1/t
        # Don't unmask anywhere already unmasked
        changes = torch.bitwise_xor(changes, \
            torch.bitwise_and(changes, unmasked))
        # Update unmasked
        unmasked = torch.bitwise_or(unmasked, changes)

        # Sample from p(c_{t-1} | c_t, Z)
        logits = Transformer(torch.cat(Z, c_t))
        dist = Categorical(logits)
        c_0_hat = dist.sample()
        c_t[changes] = c_0_hat[changes]

    # Decompress 3D latents with 3D VQ-Decoder
    imgs_3d = vqae_2d.decoder(c_t)
    return imgs_3d
```

(b) Python-like code snippet for 2D to 3D translation using our approach.


\end{document}